\documentclass[preprint,12pt]{elsarticle}

\usepackage{graphicx}
\usepackage{amssymb}
\usepackage{lineno}
\usepackage{amsmath}
\usepackage{times}
\usepackage{epsfig}
\usepackage{mathrsfs}
\usepackage[table,xcdraw]{xcolor}
\usepackage[colorlinks,linkcolor=blue]{hyperref}
\usepackage{url}
\usepackage{siunitx}
\usepackage{booktabs}
\usepackage{tabularx}

\usepackage{algorithm}
\usepackage{algpseudocode}
\usepackage{amsmath}
\journal{a Journal}

\begin{document}

\begin{frontmatter}

\title{Point Attention Network for Semantic Segmentation of 3D Point Clouds}

\author{Mingtao Feng\textsuperscript{a}, Liang Zhang\textsuperscript{b}, Xuefei Lin\textsuperscript{c}, Syed Zulqarnain Gilani\textsuperscript{d} and Ajmal Mian\textsuperscript{d}}

\address{\textsuperscript{a}Hunan University, Changsha, China; \textsuperscript{b}Xidian University, Xi'an, China; \textsuperscript{c}Hunan Agricultural University, Changsha, China; \textsuperscript{d}The University of Western Australia, Perth, Australia}

\begin{abstract}
Convolutional Neural Networks (CNNs) have performed extremely well on data represented by regularly arranged grids such as images. However, directly leveraging the classic convolution kernels or parameter sharing mechanisms on sparse 3D point clouds is inefficient due to their irregular and unordered nature. We propose a point attention network that learns rich local shape features and their contextual correlations for 3D point cloud semantic segmentation. Since the geometric distribution of the neighboring points is invariant to the point ordering, we propose a Local Attention-Edge Convolution (LAE-Conv) to construct a local graph based on the neighborhood points searched in multi-directions. We assign attention coefficients to each edge and then aggregate the point features as a weighted sum of its neighbors. The learned LAE-Conv layer features are then given to a point-wise spatial attention module to generate an interdependency matrix of all points regardless of their distances, which captures long-range spatial contextual features contributing to more precise semantic information. The proposed point attention network consists of an encoder and decoder which, together with the LAE-Conv layers and the point-wise spatial attention modules, make it an end-to-end trainable network for predicting dense labels for 3D point cloud segmentation. Experiments on challenging benchmarks of 3D point clouds show that our algorithm can perform at par or better than the existing state of the art methods.
\end{abstract}


\end{frontmatter}

%\linenumbers

%%%%%%%%%%%%%%%%%%%%%%%%%%%%%%%%%%%%%%%%% Introduction %%%%%%%%%%%%%%%%%%%%%%%%%%%%%%%%%%%%%%%%%%%%%%%%%%%%%%%%%%%%%%%%%%%%%%%%
\section{Introduction}
With the widespread availability of 3D scanning devices and depth sensors~\cite{LF_Feng}, 3D geometric data is being increasingly used in many different application domains such as robotics, autonomous driving, 3D scene understanding, city planning, infrastructure maintenance etc~\cite{PR_1,PR_2,PR_3,PR_4}. 
Several representations of 3D shape have been investigated, such as depth maps, voxels, multi-views, meshes and point clouds~\cite{Voxsegnet}. However, point cloud is arguably the simplest format for 3D data representation and has hence attracted increasing research interest. Similar to the pixels in a 2D image, points in the three-dimensional coordinate system are basic building units of point clouds, which naturally encode the geometric features and their spatial distributions of a real 3D scene.  

The extraction of meaningful information from 3D point clouds requires semantic segmentation. Point cloud semantic segmentation has been a challenging and active research topic for the last few years. Unlike pixels of 2D images which have a rectangular grid-like structure with no missing bits, 3D point clouds are sparse, irregular, unordered and with missing regions due to the limited range of scanners and occlusions. While deep learning has been very successful in semantic segmentation of 2D images, its use for 3D point clouds has not been fully exploited yet. Qi et al.~\cite{Pointnet} first proposed PointNet that learns point features directly from unordered point sets. In PointNet, all 3D points are independently passed through a set of multi-layer perceptions (MLP) and then aggregated to a global feature using max-pooling. Recent research directions focus on extending the basic idea of PointNet to incorporate local geometric features for abstracting more discriminative high level features~\cite{Pointnet++,DynamicEdge,ECC}. Among these methods, Pointnet++ \cite{Pointnet++} exploited neighborhood points within a ball query radius, where each local point is processed separately by a PointNet-based hierarchical network. However, the relationships between local points are neglected. Recently dynamic graph CNN \cite{DynamicEdge} was proposed which considers neighborhood points as a local graph and uses a filter generating network to assign edge labels. Since the edge-conditioned network does not consider the order of local points, it does not have transformation invariance. Similar to dynamic graph CNN~\cite{DynamicEdge}, dynamic edge conditioned filters \cite{ECC} were introduced as an edge function to encode local information by combining the relative coordinates (raw features) between the center point and its $K$-nearest neighbors (KNN). Although dynamic edge conditioned filters~\cite{ECC} attempt to use a function designed to handle local points, it does not fully exploit the geometrical correlations of the local neighborhood points. 

To address the above short comings, we propose a local attention-edge convolution (LEA-Conv) layer that extends the ideas of ~\cite{Pointnet++, DynamicEdge} and~\cite{ECC}. The LAE-Conv layer constructs a local graph based on the neighborhood points searched along multiple directions. Unlike KNN and ball query methods, we propose a multi-directional search strategy that finds all neighborhood points from 16 directions spread systematically within a ball query making the local geometric shape more generalizable across space. After the search operation, LAE-Conv layer assigns attention coefficients to each edge and then aggregates the central point features as a weighted sum of its neighbors. Aggregating features from a group of points with their contribution coefficients, rather than a single max-pooling operation, better exploits the correlations between points to get accurate and robust local geometric details. Moreover, LAE-Conv layer is invariant to the ordering of points and can implicitly infer how the points contribute to the overall 3D shape.

Equipped with the LAE-Conv layer, we are able to design hierarchical deep learning architectures on point clouds for semantic segmentation. Since each LAE-Conv layer has a limited local receptive field, each unit of the output features (at the initial layers) exploits correlations within its local scale only. However, later LAE-Conv layers have progressively larger receptive fields enabling the network to learn hierarchical features. While existing networks~\cite{Pointnet++,PointCNN,DynamicEdge} capture multi-scale shapes for high-level point feature learning, they do not leverage the long-range contextual relationship among points belonging to the same categories, which is important for semantic segmentation. Superpoint graphs \cite{Large_superCVPR18} employed a recurrent neural network to exploit long-range dependencies based on an unsupervised geometric partitioning. However, that method relies heavily on the partitioning results. To address the above problems, in this paper, we propose a point-wise spatial attention module, which captures long-range contextual information in the spatial dimension. Features obtained from LAE-Conv layer are fed into the point-wise spatial attention module to generate a global dependency matrix which models the correlations between any two points of the feature maps. Through multiplying the dependency matrix with original features, the differences between point features of the same category are reduced. Hence, any two points with similar features can contribute mutual improvement regardless of their spatial distance.

Using the proposed LAE-Conv layer and point-wise spatial attention model as the main building blocks, we design a U-shape network to predict the dense labels for semantic segmentation of 3D point clouds. The unorganized 3D points (raw data) are input directly to our point attention network comprising an encoder and a decoder. This is different from other approaches~\cite{Pointnet++,PointCNN,DynamicEdge} since our method stacks the point-wise attention module after the LAE-Conv layer at different stages of the network enabling it to learn more accurate local geometric features and long range relationships.

To summarize, our contributions include: (1) A novel local attention-edge convolution (LAE-Conv) layer to encode point features using a weighted sum of its neighborhood points with edge attention coefficients. The proposed multi-directional search strategy makes the local geometric shape more generalizable across space. (2) A novel point-wise spatial attention module that learns the long-range contextual information and significantly improves the segmentation results by boosting the representation power of local features obtained from the LAE-Conv layers. (3) Extending the U-shaped network to incorporate the proposed LAE-Conv layer and point-wise spatial attention module. Experimental results show that our method obtains on pair or better performance than existing state-of-the-art methods quantitatively and qualitatively on challenging benchmark datasets. Finally, we show that our proposed point  attention  block can generalize to other networks and improve their performance.

%%%%%%%%%%%%%%%%%%%%%%%%%%%%%%%%%%%%%%%%%% Related Work %%%%%%%%%%%%%%%%%%%%%%%%%%%%%%%%%%%%%%%%%%%%%%%%%%%%%%%%%%%%%%%%%%%%%%%%
\section{Related Work}

A number of deep learning architectures have been recently proposed to learn directly from 3D point cloud data or its derived representations for applications such as semantic segmentation, object part segmentation and object categorization. We provide a brief survey of these methods and divide them into three categories based on the underlying data representations they use.

\subsection{Indirect methods} This category includes methods that transform the irregular 3D point cloud data to a canonical form so that traditional convolutions can be applied~\cite{PointGrid,SpiderCNN}. Volumetric representations~\cite{Voxel_1,Voxsegnet,Voxel_3,Voxel_4,VoxelNet} are the most common canonical form used by these methods due to their simplicity. However, voxel representations have cubic complexity leading to dramatic increase in the memory consumption and computing resources required to process even medium size point clouds. To alleviate this problem, Octree-Net~\cite{O-cnn,Octnet_1} and Kd-Net~\cite{KdNet} have been proposed which skip representation and computations at empty spaces to save memory and processing resources respectively~\cite{Spherical}. Moreover, sparse convolutional operations, where the activations are kept sparse in the convolution layers \cite{Submanifold,SBnet},  have been introduced to process spatially-sparse 3D point clouds. Nevertheless, the kernels are still dense and inefficient in their implementation. Multi-view convolutional neural networks and their variants~\cite{Point_depth,3DMV_point,PVNet_point,Muti_point} have also been proposed. These methods render the 3D shape from multiple pre-defined views, which are then processed by conventional image-based convolution networks. The main drawback of the multi-view frameworks is that the 3D geometric information is not always fully retained in the 2D projections.

The sparse lattice networks proposed by Hang et al.~\cite{Splatnet} project the input 3D points onto a high dimensional lattice, perform standard spatial convolution on it and then filter the features back to the input points. Matan et al.~\cite{Extension_point} extended the function over point cloud to a volumetric function, where volumetric convolution is applied and then a restriction operator is used to do the inverse action. Qiangui et al.~\cite{Recurrent_Slice} used a slice pooling layer to project unordered point clouds into an ordered format, making it feasible to apply traditional deep learning algorithms. Fully convolutional networks \cite{FC_point} have been proposed that sample the input point cloud uniformly and use PointNet as a low-level feature learner, followed by 3D convolutions to learn features at multiple scales. Finally, tangent convolutions \cite{Tangent_3D} have also been proposed that operate directly on surface geometry in the tangent space. Although the above methods have used deep learning techniques to realize the 3D data analysis tasks, they have not used the 3D point clouds directly. We believe that learning directly from raw 3D point cloud data can achieve higher accuracy and efficiency as learning from raw data is the major strength of deep learning.

\subsection{Graph convolution methods} Graph convolutional methods combine the power of convolution operation with graph representations of irregular data. Graph convolutional networks have been designed to perform convolutions either in the spectral or spatial domain. More recently, Joan et al.~\cite{Spectral_graph} proposed a generalization of convolution for graph via the Laplacian operator. In that method, the spectral network can learn convolutional layers with a number of parameters for low dimensional graphs. Wang et al.~\cite{Local_Spectral} proposed a local spectral graph convolution to construct local graph from a point's neighborhood and aggregate information from nodes using their spectral coordinates. The PointNet++ architecture is then applied along with the local spectral graph convolution layers and graph pooling layers. The regularized graph convolution network proposed by Gusi et al.~\cite{RGCNN_graph} treats point cloud as a graph and defines convolution operation over it. Moreover, a graph smoothness prior is used in the loss function to regularize the learning process. Graph Laplacian based methods have a number of drawbacks including the computational complexity of Laplacian eigen-decomposition, the large number of parameters to express the convolutional filters, and the lack of spatial localization. Different from these methods, Martin et al.~\cite{ECC} proposed a convolution-like operation on graph signals in the spatial domain and used an asymmetric edge function to describe the relationships between local points. However, the edge labels are dynamically generated and hence, the irregular distribution of local points is not taken into account. This method was improved by Wang et al.~\cite{DynamicEdge} through max pooling operation on local features. However, max pooling operation is still unable to fully utilize the correlations of local points. Our proposed method exploits local feature learning using a completely different approach. We propose a local attention-edge convolution layer that  learns local relationships between points.

\subsection{Point cloud methods} Many researchers have proposed deep learning architectures that learn directly from point clouds. One of the earliest methods in this category is the PointNet~\cite{Pointnet} that operates on point clouds using multi-layer perception (MLP). PointNet is robust to the global transformation of 3D shape because the spatial transformer network~\cite{T-Net} is used to learn the 3D alignment. The main limitation of PointNet is that it only relies on the max-pooling layer to learn global features. Since PointNet does not consider local relationships, Qi et al.~\cite{Pointnet++} introduced an improved network named PointNet++, which exploits local geometric features in point sets and aggregates them for hierarchical inference. However, PointNet++ still treats points within local regions individually and does not consider relationships between the neighborhood points. 

Later, Francis et al.~\cite{Multi_context_ICCV17} designed a multi-scale architecture to enlarge the receptive field over the 3D scene by incorporating larger-scale spatial grid blocks into PointNet. Loic et al.~\cite{Large_superCVPR18} used an unsupervised method to cluster input points into superpoint graphs, then fed the graphs to PointNet-based gated recurrent unit. Li et al.~\cite{PointCNN} proposed X-Conv layer instead of MLP to permute unordered local points into a latent potentially canonical order. A similar approach was proposed in \cite{Mining_Kernel}, where kernel correlation was introduced to incorporate local information extracted from point cloud by PointNet. Wang et al.~\cite{SGPN_instance} introduced a similarity group proposal network for point cloud instance segmentation, which use a similarity matrix to produce a grouping proposal based features extracted from PointNet. Different from these PointNet-based frameworks, Hua et al.~\cite{PointWise_CVPR18} presented a point-wise convolution operator that can be applied to each point of the point set. Recently, Zhao et al.~\cite{PointWeb} proposed PointWeb for point cloud processing, which connects all points densely in a local neighborhood for better encoding local geometric features. Wu et al.~\cite{PointConv} introduced PointConv, a nonlinear function kernel for point cloud, which is used to learn the translation-invariant
and permutation-invariant features in 3D space. Wang et al.~\cite{Graph_AC} designed a graph attention kernel to adapt to the local geometric, which is useful for fine-grained segmentation.

A common limitation of all the aforementioned methods is that they are unable to simultaneously exploit fine local details and long-range contextual information. We fill this gap and propose a network that learns local geometrical features using their edge attention coefficients and allows deep learning architectures to exploit fine details as well as interactions over longer distances. 

%%%%%%%%%%%%%%%%%%%%%%%%%%%%%%%%%%%%%%%%%%%%%%%%%%% Our Approach %%%%%%%%%%%%%%%%%%%%%%%%%%%%%%%%%%%%%%%%%%%%%%%%%%%%%%%%%%%%%%
\section{Proposed Approach}

\begin{algorithm}[t] %%%%%%%%%%%%%% Algorithm. 1  %%%%%%%%%%%%%%%%
	\caption{LAE-Conv Operation}
	\label{LAE_layer}
	\begin{algorithmic}[1]
		\Require
		Input local points $h$, central point $p_i$; Number of selected points $m$ in each bin;
		\Ensure
		Filtered central point $p_{i_{LAE}}$;
		\State Search $K$ neighbor points $p_j$ of $p_i$ in the point cloud $h$;
		\State $p_j-p_i$: move points $p_j$ to local coordinate system of $p_i$;
		\State $W(p_j-p_i)$: transform the input points into higher-level features ;
		\State $\alpha_{ij}$: compute normalized attention edge coefficients with softmax;
		\State $p_i'$: use graph attention aggregator to obtain updated feature at $p_i$;
		\State MLP$(p_i')$: feature transformation operation;\\
		\Return $p_{i_{LAE}}$;
	\end{algorithmic}
\end{algorithm}

We first give details of the LAE-Conv layer that captures accurate local geometric details. Next, we explain the point wise spatial attention module that aggregates the long-range contextual information based on the output of LAE-Conv layers. Finally, we present a general framework of our network.
% [this is in the next section] and elaborate on the main differences of our method from existing approaches.

%%%------------------------------- Local attention-edge convolution ----------------------------------%%%
\subsection{Local Attention-Edge Convolution (LAE-Conv)}\label{LAE-Conv_subsection}

The Local Attention-Edge Convolution (LAE-Conv) layer forms the basic component of our point attention network architecture for 3D point cloud semantic segmentation. Inspired by DGCNN~\cite{DynamicEdge}, ECC~\cite{ECC}, GATs~\cite{GAT_ICLR2018} and Non-local network~\cite{Non_local}, we construct a multi-directional neighborhood graph and apply graph attention mechanism to compute local edge features. Similar to traditional convolution in images, LAE-Conv explores local regions to leverage correlations between unordered points and exploits the local geometric structure of the points. We summarize the LAE-Conv operator in Algorithm~\ref{LAE_layer}.

\subsubsection{Multi-directional Search} 

\begin{figure}  %%%%%%%%%%%%%%%% Figure. 1  %%%%%%%%%%%%%%%%
	\center{\includegraphics[width=0.5\textwidth]{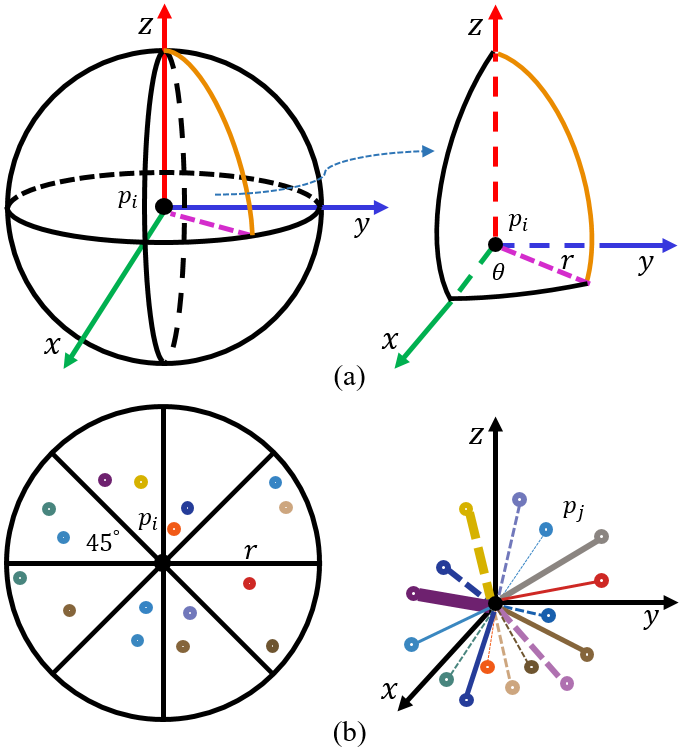}}
	\caption{\label{Neighbor_search}(a) An illustration of our multi-directional search method. The ball space around the center point within the search radius is divided into 16 uniform directions. The azimuth $\theta$, radius $r$ and number $m$ of selected points in one cube are hyperparameters. (b) When $m=1$, 16 neighborhood points are considered along different directions. If all neighbors are projected onto the $xy$ coordinate plane, we can see that there are two points in each of the eight directions. The thickness of the line connecting the center point to the neighbors represents different contributing values.}	
\end{figure}

In image convolution operation, the local region of a pixel can be represented in a grid-like structure given a convolution kernel size. However, the neighborhood of a center point (in a point cloud) is defined by metric distance in a 3D coordinate system where neighboring points are irregularly distributed. To robustly leverage local point correlations, we endeavour to explicitly capture geometric information in different orientations. Given an unordered point cloud $P=\left\{p_1,p_2,\ldots,p_N\right\}$ with $p_i\in\mathbb{R}^C$, where $N$ is the number of points, and $C$ is the feature dimension at each point. When each point is represented by its 3D coordinates $p_i=(x_i,y_i,z_i)$, then $C=3$. We denote a central point in $P$ as $p_i$, and its $K$ neighbors in $P$ as $p_j$, $j\in\mathcal{N}(i)$. As shown in Figure~\ref{Neighbor_search}(a), the space around the reference point $p_i$ within a radius of $r$ is split into 16 bins, where each bin indicates a direction. Each bin has an azimuth angle $\theta=\angle{45}$. Within the spatial range represented by each bin, we select $m$ nearest points of $p_i$ from all the points that fall in that bin and use their features to represent the bin, i.e. when $m=1,2,3...$, $K=16,32,48...$. Since some points far away from $p_i$ are not very useful to represent $p_i$, we set the radius $r$ empirically as a hyper-parameter according to each layer. In case there are insufficient points inside a bin, point $p_i$ is repeated. This is similar to self convolution.

Two common ways for range query are K-nearest neighbor (KNN) search and ball query. KNN returns a fixed number of $K$ neighboring points while ball query returns all points that are within a radius. The local shape will not be well represented if all selected points, using either of the methods, are from a small region or one direction. Different from KNN and ball query, our search method guarantees that neighborhood points are from different directions to ensure sufficient expressive power of encoding the local geometric information. We compare the effectiveness of our search method over ball query and KNN in the experiments section.

\subsubsection{Aggregation} For a set of local points $h=\left\{p_i,p_{j_1},p_{j_2},\cdots p_{j_K} \right\}$, $h\in\mathbb{R}^C$, where $p_i$ is the central point and others are its $K$ neighbors, we consider a graph $G=(V,E)$, where $V$ is a finite set of points with $|V|=K+1$ and $E \subseteq V \times V$ is a set of directed edges $\left\{ (p_i, p_{j_1}), (p_i, p_{j_2})\cdots(p_i, p_{j_{K}}) \right \}$. We define the attention edge coefficients as $e_{ij}$, which represent the importance of neighbors $p_j$ to the central point $p_i$, computed by an attention mechanism $a$.

\begin{equation}
e_{ij}=a(Wh_i,Wh_j) \label{Edge_coef}
\end{equation}

Where $W\in\mathbb{R}^{C \times C'}$ is a learnable weight matrix that transforms the input point set to higher-level features, $h_i$ and $h_j$ represent the central point and its neighbors respectively and the mechanism $a$ is a single layer MLP, parametrized by a weight vector $\Vec{\mathrm{a}}\in\mathbb{R}^{C'}$. To make the edge coefficients easily comparable across different points, we use the softmax function to normalize them across all neighbors of the reference point $p_i$:

\begin{equation}
\alpha=\text{softmax}(e_{ij})=\frac{\exp{(e_{ij})}}{\sum_{j\in\mathcal{N}(i)}\exp{(e_{ij})}}. \label{Normalization_coef}
\end{equation}

The final edge coefficients computed by the attention mechanism may then be expressed as:

\begin{equation}
\alpha_{ij}=\frac{\exp{(a(W(p_j-p_i)))}}{\sum_{j\in\mathcal{N}(i)}\exp{(a(W(p_j-p_i)))}},
\end{equation}

\noindent Where the neighbor points of the central point are transformed to local coordinate systems by  $(p_i-p_j)$ and then the local coordinates of each point are lifted to higher-order features by $W$.

Once obtained, the normalized edge coefficients $\alpha_{ij}$ are used to assign attributes to each edge. Our approach computes the filtered feature at point $p_i$ as a weighted sum of points in its neighborhood. The proposed commutative aggregation method not only solves the problem of undefined point ordering, but also smoothes out the structural information. The local graph attention aggregator is defined as

\begin{equation}
p_i'=\sum_{j\in\mathcal{N}_{p_i}} \alpha_{ij}Wp_j ,
\end{equation}

\noindent where $p_i'$ is the updated features of central point $p_i$.

\subsubsection{Transformation} Now we have an aggregated representation for the central point $p_i$. It is natural to add a feature transformation function $f$ to incorporate additional non-linearity and increase the learning capacity of the model. The transformation can be realized by MLP with a non-linear activation function. The output of the transformation function is $p_{i_{LAE}}:1\times C'$. The proposed LAE-Conv layer is described in Algorithm~\ref{LAE_layer}.

%%----------------------------- Point-wise spatial attention model --------------------------------------%%
\subsection{Point-wise Spatial Attention Block}

The output point cloud $P_{LAE}$: $N \times C'$ of the LAE-Conv layer have rich representation power for local geometric features. However, since each LAE-Conv layer have a local receptive field, individual units of the filtered features are unable to exploit contextual information outside of their local regions. In $P_{LAE}$, features corresponding to the points with the same label are significantly different when the points are far apart. These differences affect the point wise segmentation accuracy of the scene as a whole. To address this issue, we focus on the global spatial relationships to boost the representation power of the LAE-Conv layer. We design a point-wise spatial attention module that captures the global dependencies by building associations among features within the point set. We demonstrate that by stacking these blocks after LAE-Conv layers, we can construct local-global architectures that adaptively encode long-range contextual information, thus improving the semantic segmentation accuracy of 3D point clouds that cover large areas. Next, we introduce a process to adaptively aggregate point-wise spatial contexts.

\begin{figure*}  %%%%%%%%%%%%%% Figure. 2  %%%%%%%%%%%%%%%%
	\center{\includegraphics[width=1\textwidth]{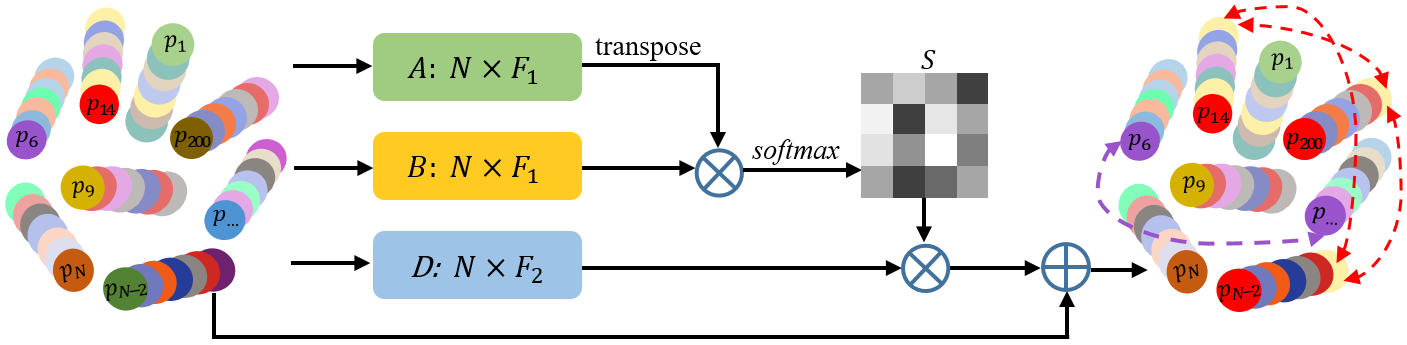}}
	\caption{\label{Long_relation} The proposed point-wise spatial attention module. The feature maps are represented by the shape of their tensors, e.g., $N\times F_1$ where $N$ denotes the number of points and $F_1$ denotes the feature dimension. For simplicity, we set the batch size to 1. $\otimes$ denotes matrix multiplication, and $\oplus$ denotes point-wise sum. The green, yellow and blue boxes denote MLP layers. Long-range correlations are learned once the input features pass through this module. }
	\vspace{-3mm}
\end{figure*}

Inspired by the position attention operation~\cite{Dual_attention}, we define a point-wise spatial attention module for 3D point clouds. As illustrated in Figure~\ref{Long_relation}, two MLP layers are used to transform the local feature $P_{LAE}$ into two new representations $A$ and $B$ respectively, where ${A,B}\in\mathbb{R}^{F_1}$. We compute relationships between different points based on the transpose of $A$ and $B$. Unlike~\cite{Dual_attention}, we calculate the spatial correlations of all points directly from the transpose of $A$ and $B$ without reshaping the matrices, hence, maintaining the original space distribution. Softmax is then applied to normalize relationship map to get the point-wise spatial attention map $S$ with size $N \times N$: 

\begin{equation}
s_{ij}=softmax\left(\frac{exp(A_i \cdot B_j)}{\sum_{i=1}^N exp(A_i \cdot B_j)}\right) ,
\end{equation}

\noindent where $i$ and $j$ denote the point positions in $A$ and $B$ respectively, $s_{ij}$ is the $i^{th}$ point's impact on the $j^{th}$ point, and $\cdot$ denotes matrix multiplication. We show that two points have a strong correlation when their features have similar semantic information.

At the same time, the local feature $P_{LAE}$ is transformed to a new feature $D\in\mathbb{R}^{F_2}$ by an MLP layer. This is followed by a matrix multiplication between $S$ and $D$. Finally, the output is multiplied by a scale parameter $\alpha$ and element-wise summation is performed with the features $P_{LAE}$ to obtain the final output $P_{final}\in \mathbb{R}^{N \times C''}$ as follows:

\begin{equation}
P_{final}= S\cdot D + P_{LAE} ,
\end{equation}

\noindent where $\cdot$ denotes matrix multiplication. Here, the resulting feature $P_{final}$ contains a long-range contextual information and selectively aggregates contexts according to the point-wise spatial attention map $S$. This module improves the feature representation power and is more accurate for 3D point cloud semantic segmentation. 

\begin{figure*}  %%%%%%%%%%%%%%%%%%%%%%%% Figure. 3 %%%%%%%%%%%%%%%%
	\center{\includegraphics[width=\textwidth]{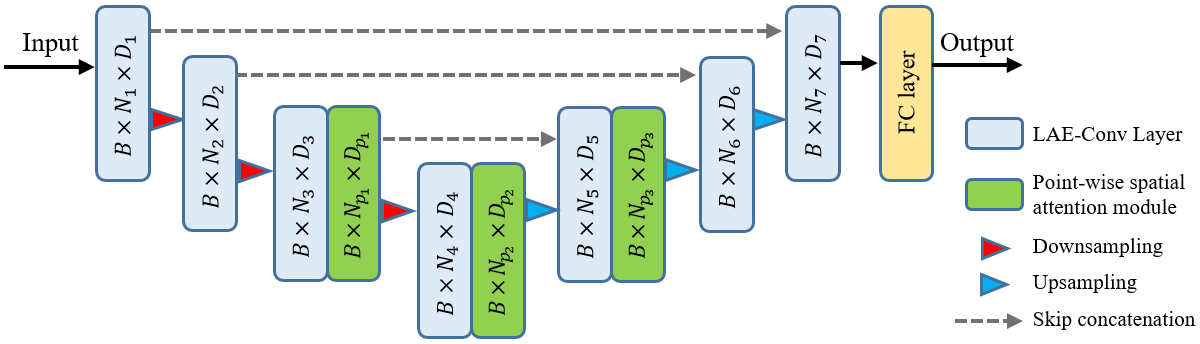}}
	\vspace{-3mm}
	\caption{\label{Network_architecture} Illustration of the proposed point attention network for point cloud segmentation. The encoder and decoder parts are based on the LAE-Conv layer and point-wise spatial attention module. $B$, $Ni$ and $Ci$ denote the batch size, point number and point feature dimension respectively. The downsampling and upsampling processes are followed by~\cite{Pointnet++}. The encoder and decoder parts are linked by three skip connections. }
	\vspace{-3mm}
\end{figure*}

%%--------------------------------- network architecture ----------------------------------------%%
\subsection{Network Architecture}
For dense point label prediction, the output resolution is high. Moreover, there are multiple objects with different scales in one scene. Selecting the most representative scale for each kind of object is important for semantic segmentation. Following the hierarchical structure of PointNet++~\cite{Pointnet++}, our network consists of encoder and decoder parts. As shown in Figure~\ref{Network_architecture}, our point attention network comprises the LAE-Conv layers and point-wise spatial attention modules. At the encoder part, the input point set is processed by three LAE-Conv layers, which transform it into fewer representation points but with richer features. The input point cloud is represented by its 3D coordinates and sometimes with the RGB color values as well. The point-wise spatial attention modules are stacked after the third and fourth LAE-Conv layers to aggregate long range point-wise contextual information from output of the previous LAE-Conv layer. The long-range contextual features along with the local features from LAE-Conv layers together achieve robust and accurate 3D point cloud semantic segmentation.

At the decoder part, three skip connections are used to combine features from the encoders. The point-wise spatial attention module is also inserted after the fifth LAE-Conv layer at the decoder part. In our hierarchical architecture, we use three steps of down-sampling operations and tree steps of up-sampling operations which are followed by set abstraction and feature propagation modules as in PointNet++~\cite{Pointnet++}. Finally, all the features in the last decoder layer go through fully connected layer and convert to class probabilities.

%%%%%%%%%%%%%%%%%%%%%%%%%%%%%%%%%%%%%%%%%%%%%%%%%%% Compared to Existing Methods %%%%%%%%%%%%%%%%%%%%%%%%%%%%%%%%%%%%%%%%%%%%%%%%%%%%%%%%%%%%%%
\section{Comparison with Existing Methods}
Our point attention network is a more generalized form of the classic approach PointNet++~\cite{Pointnet++}. We explain how PointNet++ is a special case of our network. PointNet++ is an extension of~\cite{Pointnet} with considers local point structure. Given a reference point $p_i$, ball query search $K$ local points with data size $N_l \times K \times C_l$, PointNet processes the local region points individually and then max pools them to get the most representative point feature as the output $N_l \times C_l'$ of the local region. Different from PointNet++, the LAE-Conv layer constructs the local graph for the $K$ neighbors and central point $p_i$. We compute attention edge coefficients $e_{ij}={e_{i1},e_{i2},\cdots e_{iK}}$ to indicate different contributions of each neighbor to the central point. When $e_{ij}=\left\{e_{i1}=1|e_{i2},e_{i2},\cdots e_{iK}=0\right\}$, $p_1$ features are selected to represent the local region. We can observe that the basic convolution layer of PointNet++ is an instance of our LAE-Conv layer. 

DGCNN~\cite{DynamicEdge} uses KNN to establish local point shape and proposes an aggregation operation $max(MLP(p_i,p_j-p_i))$. In that operation, the neighbor points are moved to the local coordinate system first and then stacked with the central point. All the neighbors have equal contribution to the central point, which is equivalent to our operator when all edge coefficients are equal to $1$. Since DGCNN is based on PointNet, the receptive field remains constant ($K$) at different layers, which is a disadvantage when encoding point clouds with different spatial distribution densities.

Similar to PointNet++, PointCNN~\cite{PointCNN} follows the encoder-decoder architecture and learns a $\mathcal{X}$ transformation to lift the input irregular points into an unknown canonical format, then applying a typical convolution on the transformed point cloud. In PointCNN, the dilated convolution process from image convolution networks is employed to expand the local receptive field of different layers. The local receptive field changes the number of neighborhood points $K$ by adjusting the dilation ratio. Different from the grid structure of local pixels, points are disordered in a three-dimensional coordinate system and the density distribution is not uniform. Although KNN searches for neighborhood points which is controlled by the dilation ratio proportionally, the global geometric features learned by the change of receptive field is limited. To address this issue, our point attention network inserts a point-wise attention module in the high level feature layer. A crucial difference between these two operations is that the latter assumes a long range dependency, which reduces the gap between features corresponding to the points with the same label encoding more accurate global information. The more similar are the feature representations of the two points, the greater is the correlation between them.

%%%%%%%%%%%%%%%%%%%%%%%%%%%%%%%%%%%%%%%%%%%%%%%%% Experiments and Discussion %%%%%%%%%%%%%%%%%%%%%%%%%%%%%%%%%%%%%%%%%%%%%%%%%%%%%%%%%%%%
\section{Experiments and Discussion}
We evaluate the performance of the proposed network on the  ShapeNet~\cite{ShapeNet_2} 3D part segmentation dataset and the two largest point cloud segmentation benchmarks, ScanNet~\cite{Scannet} and Stanford Large-Scale 3D Indoor Spaces (S3DIS)~\cite{S3DIS}. While ShapeNet is synthetic data, ScanNet and S3DIS are real point clouds obtained with a scanner. We perform ablation studies of different design choices and network variations as well as compare the performance of our network with existing state of the art.

%%--------------------------------------------- ScanNet -------------------------------------------------%%
\subsection{ScanNet}

ScanNet~\cite{Scannet} contains 1513 scans annotated with semantic voxel labels from 21 categories (bed, refrigerator, floor, table etc. plus other furniture). ScanNet is divided into 1201 training and 312 test samples. Similar to~\cite{Pointnet++}, we split the ScanNet training scenes into 2m by 2m by 3m blocks, with 0.5m padding in each direction ($x$,$y$,$z$) and sample 8192 points randomly from each block on the fly. To predict semantic label of every point of the test scene, we similarly split it into similar cubes using a sliding window strategy along the $xy$ plane with different stride sizes. If the same point gets different predictions in the overlap regions, we choose the one with highest confidence.

Although ScanNet also contains RGB values for each point, we only use the $xyz$ coordinates as point features for a fair comparison with other methods. Hence, the input data size for the network is $8192\times3$. As shown in Figure~\ref{Network_architecture}, we use downsampling and upsampling operations from PointNet++~\cite{Pointnet++} for both the encoder and decoder parts. The output point numbers and feature dimensions of different LAE-Conv layers are $(N_1=8192, C_1=64)$, $(N_2=2048, C_2=128)$, $(N_3=512, C_3=256)$, $(N_4=128, C_4=512)$, $(N_5=512, C_5=256)$, $(N_6=2048, C_6=256)$ and $(N_7=8192, C_7=128)$ respectively. The fully connected layer with size $(N_{fc}=8192,C_{fc}=21)$ converts the final features into class probabilities. We set $(m=1,K=16)$ for the neighborhood search. For the three point-wise attention block, the output point numbers and feature dimensions are $(N_{p_1}=512,C_{p_1}=256)$, $(N_{p_2}=128,C_{p_2}=512)$ and $(N_{p_3}=512,C_{p_3}=256)$ respectively. The initial learning rate is 0.001, batch size is 22 and the momentum is 0.9. We set the decay rate of 0.7 and stop training after 1000 epochs. 

\begin{table} %%%%%%%%%%%%%%%%%%%%%%%%%%%%%% Table. 1 %%%%%%%%%%%%%%%% 
	\centering
	\caption{\label{Scannet_acc} 3D point cloud semantic segmentation results on ScanNet scenes. The metrics are mean per-class Intersection over Union (mIoU,\%) and per voxel overall accuracy (OA, \%).}
	\begin{tabular}{l|c|c}
		\toprule
		Method                              & mean IoU & Overall Accuracy (OA)  \\
		\midrule
		PointNet~\cite{Pointnet}            &  -   & 73.9\\ 
		PointNet++~\cite{Pointnet++}        &  -   & 84.5\\ 
		RSNet~\cite{Recurrent_Slice}        & 39.35& -   \\ 
		TCDP~\cite{Tangent_3D}              & 40.9 & 80.9\\ 
		FCPN~\cite{FC_point}                &   -  & 82.6 \\ 
		%3DMV~\cite{3DMV_point}              &      &      \\ 
		3DRCNN~\cite{3DRCNNet}              &      & 76.5 \\
		PointCNN~\cite{PointCNN}            &   -   & 85.1 \\ 
		\midrule
		Ours   & \textbf{42.1}  & \textbf{86.7}\\ 
		\bottomrule
	\end{tabular}
\end{table}

\begin{table}%%%%%%%%%%%%%%%%%%%%%%%%%%%%%% Table. New  %%%%%%%%%%%%%%%%
	\centering
	\caption{\label{Parameter_compare} Model size and inference time comparison, where "M" means million and "s" denotes second. We use the model file (.cptk) size obtained by the training using tensorflow to represent the complexity of different methods. The entire scenes was tested 5 times and the average time was recorded.}
	\begin{tabular}{l|c|c}
		\toprule		
		Methods & Size (M) & Time (s)   \\ 
		\midrule
		PointNet~\cite{Pointnet} & 321.9 & 2.16   \\
		PointNet++(msg)~\cite{Pointnet++} & 177.3 &  3.8  \\  
		DGCNN~\cite{DynamicEdge} & 180.1 & 3.94   \\ 
		SpiderCNN(3-layers)~\cite{SpiderCNN} & 349.5 & 4.3   \\		
		\midrule		
		Ours & 183 &  3.97  \\ 
		\bottomrule		
	\end{tabular}
\end{table}

\begin{figure*}  %%%%%%%%%%%%%%%%%%%%%%%%%%% Figure. 4 %%%%%%%%%%%%%%%%
	\center{\includegraphics[width=0.96\textwidth]{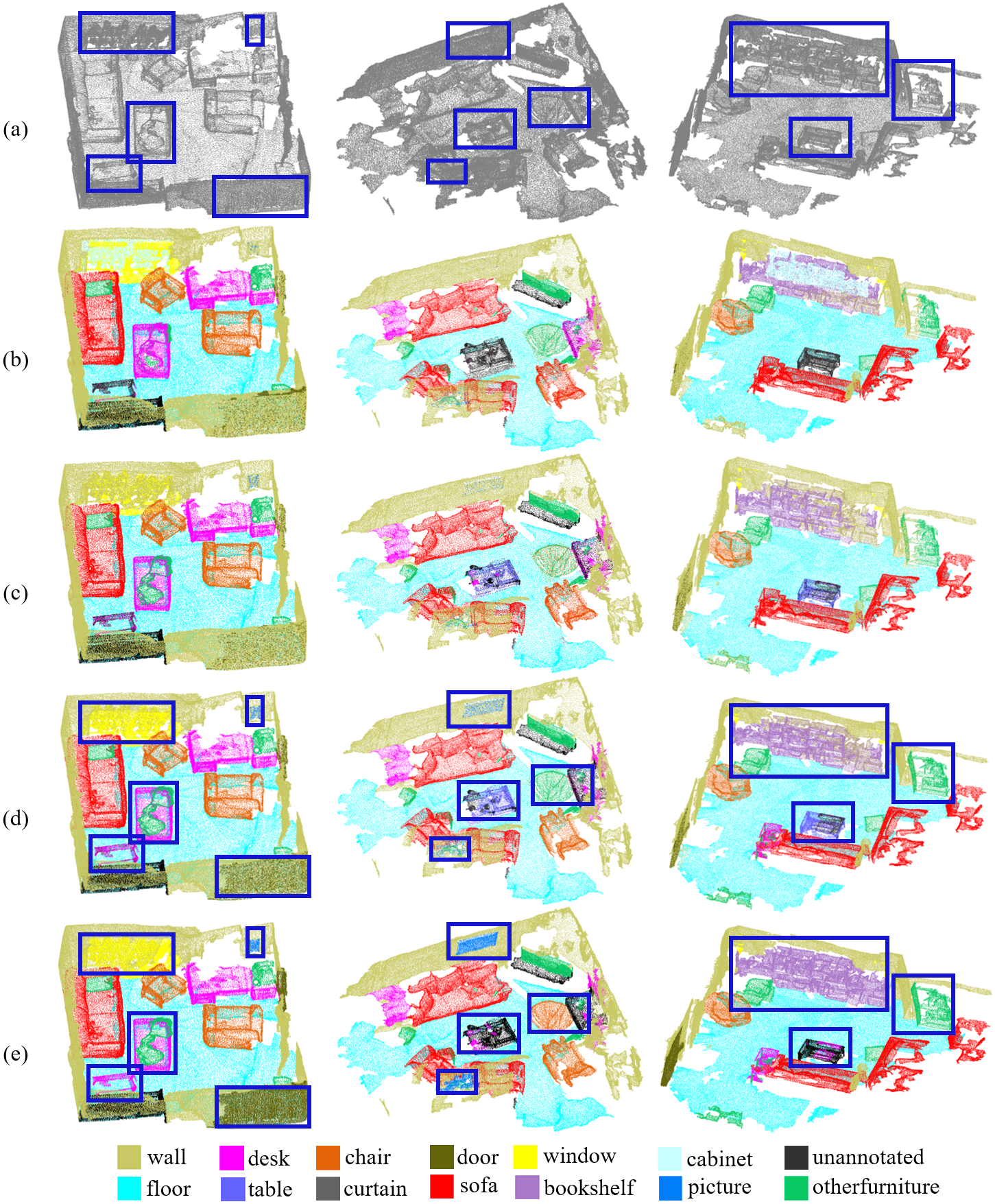}}
	%	\vspace{-1mm}
	\caption{\label{Scannet_Compare} Qualitative comparison on three scenes from ScanNet. (a) Input point cloud with only $xyz$ coordinate features. Semantic segmentation by (b) PointNet++~\cite{Pointnet++}, (c) PointCNN~\cite{PointCNN} and (d) our method. (e) Ground truth Semantic labels. Colors denote different categories. These scenes contain only 13 categories out of 21. The areas marked by the boxes are some examples where our method performed significantly better than others.  }
	\vspace{-3mm}
\end{figure*}

Table~\ref{Scannet_acc} shows quantitative comparison of our proposed point attention network with PointNet++~\cite{Pointnet++}, PointCNN~\cite{PointCNN} on the ScanNet dataset. This comparison is done using two metrics, namely the mean per-class IoU (mIoU, $\%$) and per voxel overall accuracy (OA, $\%$). For a fair comparison, Table~\ref{Scannet_acc} shows results of baseline methods reported in the original papers since the trained models are not available for testing. Compared to the baseline methods, our network achieves the  highest accuracy on both metrics. Table~\ref{Parameter_compare} reports the model size and average inference time of a few representative methods ~\cite{Pointnet}, \cite{Pointnet++}, \cite{DynamicEdge}, \cite{SpiderCNN}, where the released source codes are easy to use. Experiments are conducted by a single NVIDIA GTX TitanX GPU with tensorflow and an Intel i7-9700K@3.6 GHZ 8 cores CPU. Compared with these methods, we can see that our proposed architecture improves segmentation results with only marginal extra computation cost.

Figure~\ref{Scannet_Compare} qualitatively compares the semantic segmentation obtained by PointNet++, PointCNN and our method. We use boxes to highlight some examples where our method performed significantly better than the competitors. In the first scene, the window and the door are embedded in the wall whereas the picture is hung on the wall making the semantic segmentation a real challenge. Our method's output is more regular than that of PointNet++ and PointCNN. The table in the lower left corner is incomplete with a mere skeleton. Hence, segmentation methods like PointNet++ and PointCNN get worse results compared to our method. In the second scene, all the methods get incorrect predictions on the chair that is close to the floor as well as the irregularly shaped desks. This is because the per class samples in ScanNet dataset are unbalanced~\cite{Scannet} making existing segmentation methods fail on the rare categories. In the third scene, our method performs better on bookshelves than others. In addition, the un-annotated object in the center of scene is misidentified as table by PointCNN and our method because its shape is more like a table than an ordinary chair.

To better understand the influence of various design choices made in our network, we analysis them on ScanNet.
\subsubsection{Ablation study on parameters of LAE-Conv layer} As mentioned in Sec~\ref{LAE-Conv_subsection}, there are three options (KNN, ball query and our multi-direction searching method) for searching the neighbors of the central point. We use ScanNet as a test benchmark to compare these options. We also set different point numbers at each cube for our proposed search method. In Table~\ref{Search_method_Comp}, we can see that our method is more efficient for selecting local point shapes. When $(m=2, K=32)$ and $(m=3, K=48)$, the segmentation accuracy is greatly reduced. This is because the parameters of LAE-Conv layer will increase as the number of neighbors increase. Too many neighbors bring information redundancy, which reduces the efficiency and accuracy of the LAE-Conv layer.

\begin{table} %%%%%%%%%%%%%%%%%%%%%%%%%%%% Table. 2 %%%%%%%%%%%%%%%%
	\centering
	\caption{\label{Search_method_Comp} Ablation analysis on ScanNet with different search methods and numbers. All results are based on our LAE-Conv layer while other settings are kept constant.}
	\begin{tabular}{l|c}
		\toprule
		Neighborhood Search Method             & Overall Accuracy (OA \%) \\ 
		\hline
		KNN (K=16)                  & 85.0             \\
		Ball query (K=16)           & 85.3             \\
		Proposed Multi-direction (m=1,K=16)  & \textbf{86.7}    \\
		Proposed Multi-direction (m=2,K=32)  & 85.9             \\
		Proposed Multi-direction (m=3,K=48)  & 84.4             \\ 
		\bottomrule
	\end{tabular}
	\vspace{-3mm}
\end{table}

\subsubsection{Ablation study on point-wise spatial attention block} 
To take full advantage of the point-wise spatial attention block, we show the segmentation results with more attention blocks in the network architecture. We add 7 attention blocks (after LAE-Conv layer $1-7$ ), 5 attention blocks (after LAE-Conv layer $2-6$) and 3 attention blocks (after LAE-Conv layer $3-6$). As shown in the first part of Table~\ref{Block_place_Compare}, more point-wise spatial attention blocks do not lead to an improvement in performance. One explanation is that more attention blocks massively increase the number of parameters and the network can not find a local optimal solution within the specified training steps on ScanNet. The second part of Table~\ref{Block_place_Compare} compares same number of attention blocks added to different stages of network. The attention block is added to the right, after the LAE-Conv layer (2,4,6) and (1,4,7) respectively. We can see that the results deteriorate when the attention blocks are added to layers with lower feature dimensions. A possible explanation is that the point features do not contain enough representative semantic information when their dimensions are low, the features of the points with the same labels are significantly different, and the number of parameters of attention block will also increase from (2,4,6) to (1,4,7). Under this condition, the effectiveness of attention block is limited. Finally, we choose to add three attention blocks to the right after LAE-Conv layers (3,4,5) in Figure~\ref{Network_architecture}. We also tested adding three attention blocks to vanilla PointNet++ (without MSG and DP~\cite{Pointnet++}) at the corresponding stages as in our network. As shown in the third part of Table~\ref{Block_place_Compare}, the performance of baseline network (vanilla PointNet++~\cite{Pointnet++}) is improved by $3.4\%$. This shows that our proposed point attention block is generic and is able to improve the performance of any network architecture.  

\begin{table} %%%%%%%%%%%%%%%%%%%%%%%%%%%%% Table. 3 %%%%%%%%%%%%%%%%
	\centering
	\caption{\label{Block_place_Compare} Ablation analysis on ScanNet comparing 3, 5 and 7 point-wise spatial attention modules added to our network and comparing the results when 3 point-wise spatial attention modules are added to different stages of our network. We also test adding three attention blocks to standard PointNet++\cite{Pointnet++}.}
	\begin{tabular}{l|c}
		\toprule
		Block Position & Overall Accuracy (OA \%) \\
		\hline
		LAE-Conv layer (1-7) & 84.9 \\ 
		LAE-Conv layer (2,3,4,5,6) & 85.7  \\
		LAE-Conv layer (3,4,5) & \textbf{86.7} \\
		%\midrule
		\hline
		LAE-Conv layer (2,4,6) & 86.0 \\ 
		LAE-Conv layer (1,4,7) & 85.5 \\
		\hline
		PointNet++\cite{Pointnet++} (vanilla) baseline & 83.3\\
		PointNet++\cite{Pointnet++} (vanilla, 3-5) & 84.7 \\
		\bottomrule
	\end{tabular}
	\vspace{-3mm}
\end{table}

%%------------------------------------------ S3DIS -------------------------------------------------%%
\subsection{S3DIS}

The Stanford Large-Scale 3D Indoor Spaces (S3DIS) dataset ~\cite{S3DIS} contains 3D scans obtained with the Matterport scanners in 6 areas from three different buildings, divided into 271 individual rooms. Each point in the scene is annotated with one label from $13$ categories (ceiling, wall, beam, chair, column etc. and clutter), and is represented by its 3D coordinates, RGB features and normalized location. The S3DIS is a highly unbalanced dataset~\cite{S3DIS}, floor, wall, chair and other common furniture items being the dominant classes in the dataset while bookcase, window and beam etc. being the rare classes. To prepare the training data, rooms in S3DIS are split into blocks of $2m \times 2m$, with $0.5m$ padding on each direction ($x$,$y$). We randomly sample 4096 points from each block during training while all points are used at test time. Similar to PointNet~\cite{Pointnet}, we follow the same 6-fold cross validation strategy across 6 areas. To obtain the overall segmentation accuracy, we evaluate 6 models on their corresponding test areas and report the average results.

For comparison, we use $xyz$ coordinates and RGB information as the point features. Therefore, the input data size to the network is $4096\times6$. As shown in Figure~\ref{Network_architecture}, we use downsampling and upsampling operations from PointNet++~\cite{Pointnet++} for both encoder and decoder parts. The output point numbers and feature dimensions of different LAE-Conv layers are $(N_1=4096, C_1=64)$, $(N_2=1024, C_2=128)$, $(N_3=512, C_3=256)$, $(N_4=128, C_4=512)$, $(N_5=512, C_5=256)$, $(N_6=1024, C_6=256)$ and $(N_7=4096, C_7=128)$ respectively. The fully connected layer with size $(N_{fc}=4096,C_{fc}=13)$ converts the final features into probability of each class. We set $(m=1,K=16)$ during the neighbors search process. For the three point-wise attention modules, the output point numbers and feature dimensions are $(N_{p_1}=512,C_{p_1}=256)$, $(N_{p_2}=128,C_{p_2}=512)$ and $(N_{p_3}=512,C_{p_3}=256)$ respectively. We set the initial learning rate to 0.001, batch size to 32 and momentum to $0.9$. We set the decay rate to 0.7 and stop the training process after 1000 epochs.

\begin{figure*} %%%%%%%%%%%%%%%%%%%%%%%%%%%%% Figure 5 %%%%%%%%%%%%%%%%%%%%%
	\center{\includegraphics[width=1\textwidth]{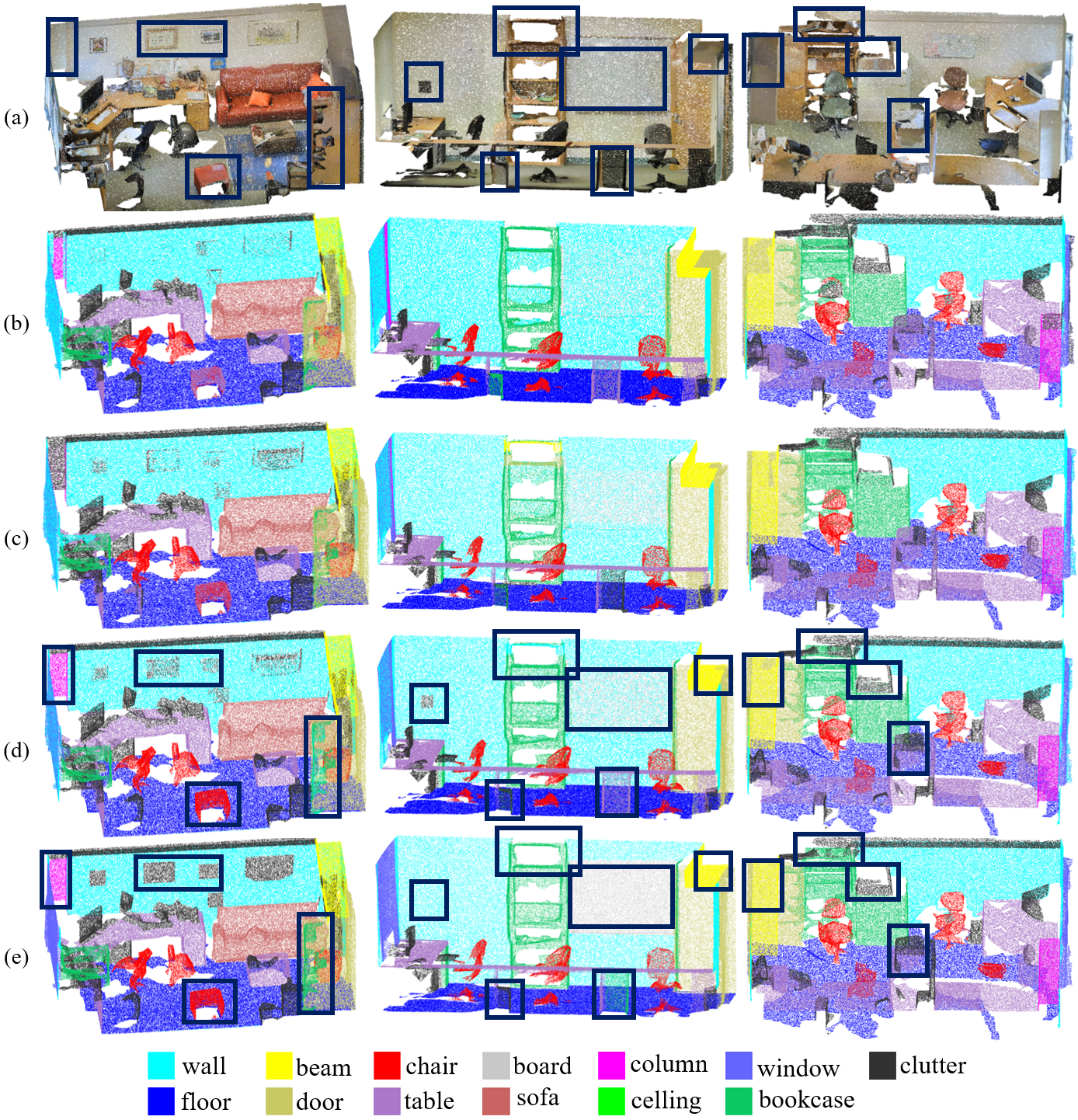}}
	\vspace{-1mm}
	\caption{\label{SI3d_Compare} Qualitative comparison on three scenes of S3DIS. (a) Input point cloud with $xyz$ coordinates and RGB features. Semantic segmentation results by (b) PointNet~\cite{Pointnet}, (c) PointCNN~\cite{PointCNN} and (d) our method. (e) Ground truth Semantic labels. Different colors denote different categories. These scenes contain 13 categories. Boxes highlight some examples where our method performs better than others.}
	\vspace{-3mm}
\end{figure*}

\begin{table*} %%%%%%%%%%%%%%%%%%%%%%%%%%%%% Table. 4 %%%%%%%%%%%%%%%%
	%\centering
	\caption{\label{S3DIS_Acc} Quantitative comparison using overall accuracy (OA,\%) and mean IoU (mIoU,\%) on S3DIS.}
	\resizebox{\textwidth}{13mm}{
		\begin{tabular}{l|c|c|ccccccccccccc}
			\toprule
			& mIoU & OA & celling & floor & wall & beam & column & window & door & chair & table & bookcase & sofa & board & clutter \\ 
			\midrule
			PointNet~\cite{Pointnet}& 47.6 & 78.5 & 88.0 & 88.7 & 69.3 & 42.4 & 23.1 & 47.5 & 51.6 & 54.1 & 42.0 & 9.6 & 38.2 & 29.4 & 35.2\\
			SPGraph~\cite{Large_superCVPR18}& 62.1 & 85.5 & 89.9 & 95.1 & 76.4 & 62.8 & 47.1 & 55.3 & \textbf{68.4} & 73.5 & 69.2 & \textbf{63.2} & 45.9 & 8.70 & 52.9\\
			RSNet~\cite{Recurrent_Slice} & 56.47 & - & 92.48 & 92.83 & \textbf{78.56} & 32.75 & 34.37 & 51.62 & 68.11 & 60.13 & 59.72 & 50.22 & 16.42 & 44.85 & 52.03\\ 
			3DRCNN~\cite{3DRCNNet} & 53.4 & 85.7 & 95.2 & \textbf{98.6} & 77.4 & 0.80 & 9.83 & 52.7 & 27.9 & \textbf{78.3} & \textbf{76.8} & 27.4 & 58.6 & 39.1 & 51.0\\
			PointCNN~\cite{PointCNN}& 65.39 & 88.14 & \textbf{94.78} & 97.3 & 75.82 & 63.25 & 51.71 & 58.38 & 57.18 & 71.63 & 69.12 & 39.08 & \textbf{61.15} & 52.19 & 58.59\\
			\midrule
			Ours   & \textbf{66.3} & \textbf{88.95} & 94.3  & 97.0 & 76.02 & \textbf{64.66} & \textbf{53.7} & \textbf{59.17} & 58.8 & 72.4 & 69.2 & 42.63 & 60.83 & \textbf{54.14} & \textbf{59.05} \\ 
			\bottomrule
	\end{tabular}}
\end{table*}

Table~\ref{S3DIS_Acc} summerizes the quantitative results where our proposed method outperforms the baseline methods PointNet~\cite{Pointnet}, SPGraph~\cite{Large_superCVPR18}, RSNet~\cite{Recurrent_Slice}, 3DRCNN~\cite{3DRCNNet} and PointCNN~\cite{PointCNN}. It is worth noting that our method achieves higher accuracy for some rare class objects, such as beam, column, window, board and clutter because our method is able to capture more global information of points that are far apart.

In Figure~\ref{SI3d_Compare}, we compare our method with PointNet~\cite{Pointnet} and PointCNN~\cite{PointCNN} qualitatively. It is not surprising that chairs are correctly segmented more often by the three baseline methods because their shapes are more consistent, they are small and not easily confused with other objects. We can see that objects such as the whiteboard hung on the wall, column and window embedded in the wall, clutter next to the table and irregular bookcases are quite difficult to segment. We also use boxes to mark some examples where our method outperformed the baseline methods. In the first scene, our method obtains more regular segmentation of the painting on the wall than PointNet~\cite{Pointnet} and PointCNN~\cite{PointCNN}. Our final segmentation result preserves the full shapes of column and bookcase next to wall while other methods mistake them for wall or clutter. We also obtain a smoother prediction for chair in the front row than the other methods. In the second scene, the board on the wall is more accurately estimated by our method compared to PointNet and PointCNN. Our method makes fewer mistakes in predicting the bookshelves, beam and clutter which are up and below the table compared to other approaches. Notably, our method can predict the clutter on the left wall, even though it is not marked by ground truth. In the third scene, our method also outputs fewer incorrect predictions for bookcase, table, chair and beam compared to other approaches.

\begin{table*}  %%%%%%%%%%%%%%%%%%%%%%%%%%%%%%%% Table. 5 %%%%%%%%%%%%%%%%
	%\centering
	\caption{\label{ShapNet_Acc} Quantitative comparison on ShapeNet part dataset~\cite{ShapeNet}. The values show  part-averaged IoU (pIoU$\%$), mean per-category pIoU (mpIoU$\%$) and per-category IoU ($\%$) scores.}
	\resizebox{\textwidth}{30mm}{
		\begin{tabular}{l|c|c|cccccccccccccccc}
			\toprule
			& pIoU & mpIoU & \begin{tabular}[c]{@{}l@{}}air\\ plane\end{tabular} & bag & cap & car & chair & \begin{tabular}[c]{@{}l@{}}ear\\ phone\end{tabular} & guitar & knife & lamp & laptop & motor & mug & pistol & rocket &\begin{tabular}[c]{@{}l@{}}skate\\ board\end{tabular}& table \\ 
			\midrule
			shapes & & &2690 & 76 & 55 & 898 & 3758 & 69 & 787 & 392 & 1547 & 451 & 202 & 184 & 283 & 66 & 152 & 5271\\
			\midrule
			PointNet~\cite{Pointnet}& 83.7 & 80.4 & 83.4 & 78.7 & 82.5 & 74.9 & 89.6 & 73.0 & 91.5 & 85.9 & 80.8 & 95.3 & 65.2 & 93.0 & 81.2 & 57.9 & 72.8 & 80.6 \\
			PointNet++~\cite{Pointnet++}& 85.1 & 81.9 & 82.4 & 79.0 & 87.7 & 77.3 & 90.8 & 71.8 & 91.0 & 85.9 & 83.7 & 95.3 & 71.6 & 94.1 & 81.3 & 58.7 & 76.4  & 82.6 \\
			DGCNN~\cite{DynamicEdge}& 85.1 & 82.3 & 84.2 & 83.7 & 84.4 & 77.1 & 90.9 & 78.5 & 91.5 & 87.3 & 82.9 & 96.0 & 67.0 & 93.3 & 82.6 & 59.7 & 75.5 & 82.0\\
			RSNet~\cite{Recurrent_Slice}& 84.9 & 81.4 & 82.7 & 86.4 & 84.1 & 78.2 & 90.4 & 69.3 & 91.4 & 87.0 & 83.5 & 95.4 & 66.0 & 92.6 & 81.8 & 56.1 & 75.8 & 82.2\\
			SGPN~\cite{SGPN_instance}& 85.8 & 82.8 & 80.4 & 78.6 & 78.8 & 71.5 & 88.6 & 78.0 & 90.9 & 83.0 & 78.8 & 95.8 & 77.8 & 93.8 & \textbf{87.4} & 60.1 & \textbf{92.3} & \textbf{89.4}\\
			ASCNet~\cite{Attentional_context}&84.6 & 81.78 & 83.8 & 80.8 & 83.5 & 79.3 & 90.5 & 69.8 & 91.7 & 86.5 & 82.9 & 96.0 & 69.2 & 93.8 & 82.5 & 62.9 & 74.4 & 80.8\\
			PCNNet~\cite{Extension_point}& 85.1 & 81.8 & 82.4 & 80.1 & 85.5 & 79.5 & 90.8 & 73.2 & 91.3 & 86.0 & 85.0 & 95.7 & 73.2 & 94.8 & 83.3 & 51.0 & 75.0 & 81.8\\
			PGrid~\cite{PointGrid}&\textbf{86.4} & 82.23 & 85.7 & 82.5 & 81.8 & 77.9 & 92.1 & 82.4 & \textbf{92.7} & 85.8 & 84.2 & 95.3 & 65.2 & 93.4 & 81.7 & 56.9 & 73.5 & 84.6\\
			SPLATNet~\cite{Splatnet}& 85.4 & 83.69 & 83.2 & 84.3 & \textbf{89.1} & 80.3 & 90.7 & 75.5 & 92.1 & 87.1 & 83.9 & \textbf{96.3} & 75.6 & 95.8 & 83.8 & 64.0 & 75.5 & 81.8\\
			KCGP~\cite{Mining_Kernel}& 84.7 & 82.21  & 82.8 & 81.5 & 86.4 & 77.6 & 90.3 & 76.8 & 91.0 & 87.2 & 84.5 & 95.5 & 69.2 & 94.4 & 81.6 & 60.1 & 75.2 & 81.3\\
			SpiderCNN~\cite{SpiderCNN}& 85.3 & 81.7& 83.5 & 81 & 87.2 & 77.5 & 90.7 & 76.8 & 91.1 & 87.3 & 83.3 & 95.8 & 70.2 & 93.5 & 82.7 & 59.7 & 75.8 & 82.8\\
			SONet~\cite{SONet}& 84.9 & 81.0 & 82.8 & 77.8 & 88.0 & 77.3 & 90.6 & 73.5 & 90.7 & 83.9 & 82.8 & 94.8 & 69.1 & 94.2 & 80.9 & 53.1 & 72.9 & 83.0\\
			SSCN~\cite{Submanifold}& 85.98 & 83.3 & \textbf{84.1} & 83.0 & 84.0 & \textbf{80.8} & \textbf{91.4} & 78.2 & 91.6 & \textbf{89.1} & 85.0 & 95.8 & 73.7 & 95.2 & 84.0 & 58.5 & 76.0 & 82.7\\
			PointCNN~\cite{PointCNN}& 86.1 & \textbf{84.6} & \textbf{84.1} &\textbf{ 86.5} & 86.0 & \textbf{80.8} & 90.6 & 79.7 & 92.3 & 88.4 & 85.3 & 96.1 & 77.2 & \textbf{95.3} & 84.2 & 64.2 & 80.0 & 83.0\\
			\hline
			Ours & 85.9 & 84.10 & 83.3 & 86.1 & 85.7 & 80.3 & 90.5 & \textbf{82.7} & 91.5 & 88.1 & \textbf{85.5} & 95.9 & \textbf{77.9} & 95.1 & 84.0 & \textbf{64.3} & 77.6 & 82.8 \\ 
			\bottomrule
	\end{tabular}}
\end{table*}

%%------------------------------------------ ShapeNet -------------------------------------------------%%
\subsection{ShapeNet}
We also extend our network architecture to perform part segmentation on the ShapeNet dataset~\cite{ShapeNet}, which consists of $16881$ shape models from 16 object categories. Each object in ShapeNet is annotated with $2$ to $5$ parts. We follow the settings from \cite{ShapeNet_2} to divide the ShapeNet dataset for training, validation and testing. During training, we randomly sample 2048 points from each 3D shape while all points from each 3D shape are used during the test stage.

For a fair comparison, we only use the $xyz$ coordinates as the point features. The size of input data for the network is $2048\times3$. The network architecture is illustrated in Figure~\ref{Network_architecture}, we adjust the network parameters to suit ShapeNet. The output point numbers and feature dimensions of different LAE-Conv layers are $(N_1=2048, C_1=64)$, $(N_2=1024, C_2=128)$, $(N_3=256, C_3=256)$, $(N_4=128, C_4=512)$, $(N_5=256, C_5=256)$, $(N_6=1024, C_6=256)$ and $(N_7=2048, C_7=128)$ respectively. A fully connected layer with size $(N_{fc}=128,C_{fc}=16)$ is used at the end to convert the point features into part predictions. We set $(m=1,K=16)$ for the neighborhood search. For the three point-wise attention block, the output point numbers and feature dimensions are $(N_{p_1}=256,C_{p_1}=256)$, $(N_{p_2}=128,C_{p_2}=512)$ and $(N_{p3}=256,C_{p_3}=256)$ respectively. We set the initial learning rate to 0.003, batch size to 16, momentum to 0.9, decay rate to 0.7 and stop the training after 500 epochs.

We use the same evaluation metric (mean IoU) on points as PointNet~\cite{Pointnet} to compare our method with others methods~\cite{Pointnet,Pointnet++,DynamicEdge,Recurrent_Slice,SGPN_instance,Attentional_context,Extension_point,PointGrid,Splatnet,Mining_Kernel,SpiderCNN,SONet,Submanifold,PointCNN}. We report the part-averaged IoU (pIoU$\%$), mean per-category pIoU (mpIoU$\%$) and per-category IoU ($\%$) scores in Table~\ref{ShapNet_Acc}. Our method achieves on par performance with most methods in the metrics pIoU and mpIoU. In individual categories, we rank the best in ear phone, lamp, motor and rocket. As we can see, our method performs better when there are fewer data points as in the case of ear phone, motor and rocket.

%%%%%%%%%%%%%%%%%%%%%%%%%%%%%%%%%%%%%% Conclusion %%%%%%%%%%%%%%%%%%%%%%%%%%%%%%%%%%%%%%%%%%%%%%%
\section{Conclusion}
We proposed a point attention network for 3D point cloud semantic segmentation. Our network adaptively integrates local point features and long-range contextual information. We introduced a novel local attention-edge convolution (LAE-Conv) layer which exploits attention mechanism on a local graph constructed by the central point and its neighborhood to capture accurate and robust geometric details. To refine the output local features of LAE-Conv layer, we proposed a point-wise spatial attention module and showed that this module can generalize to other networks to improve their accuracy. Finally, we adapted the U-shaped network to combine the LAE-Conv layer and point-wise spatial attention modules. Experiments on challenging benchmark datasets show that our method quantitatively and qualitatively obtains on pair or better performance than existing state-of-the-art in 3D point cloud semantic segmentation.

\section*{Acknowledgment}
This work was supported in part by National Natural Science Foundation of China under Grant 61573134, Grant 61973106 and in part by the Australian Research Council (ARC) grant DP190102443. Thank Yifeng Zhang and Tingting Yang from Hunan University for helping with baseline experiments setup.

\section*{Reference}

\bibliographystyle{model1-num-names}
\bibliography{Reference.bib}

\end{document}